%% file: main.tex
\documentclass{article}

\usepackage[final,nonatbib]{nips_2018}

\usepackage[utf8]{inputenc} 
\usepackage[T1]{fontenc}    
\usepackage{hyperref}       
\usepackage{url}            
\usepackage{booktabs}       
\usepackage{amsfonts}       
\usepackage{nicefrac}       
\usepackage{microtype}      
\usepackage{amsmath}
\usepackage{algorithm}
\usepackage[noend]{algpseudocode}
\usepackage[usenames, dvipsnames]{color}
\usepackage{amssymb}        
\usepackage{graphicx}
\usepackage{listings}
\usepackage{makecell}
\usepackage{placeins}

\usepackage{color}
\newcommand{\comment}[1]{\textcolor{red}{#1}}

\usepackage[
backend=biber,
style=numeric,
sorting=ynt
]{biblatex}
 
\def \O {\varnothing}

\title{Mesh-TensorFlow:\\ Deep Learning for Supercomputers}

\author{Noam Shazeer, Youlong Cheng, Niki Parmar,\\
  \textbf{Dustin Tran, Ashish Vaswani, Penporn Koanantakool, Peter Hawkins, HyoukJoong Lee} \\
  \textbf{Mingsheng Hong, Cliff Young, Ryan Sepassi, Blake Hechtman} \\
  Google Brain\\
  \texttt{\{noam, ylc, nikip, trandustin, avaswani, penporn, phawkins,}\\
  \texttt{hyouklee, hongm, cliffy, rsepassi, blakehechtman\}@google.com} \\
}

\begin{document}

\maketitle

\begin{abstract}

Batch-splitting (data-parallelism) is the dominant distributed Deep Neural Network (DNN) training strategy, due to its universal applicability and its amenability to Single-Program-Multiple-Data (SPMD) programming.  However, batch-splitting suffers from problems including the inability to train very large models (due to memory constraints), high latency, and inefficiency at small batch sizes.  All of these can be solved by more general distribution strategies (model-parallelism).  Unfortunately, efficient model-parallel algorithms tend to be complicated to discover, describe, and to implement, particularly on large clusters.  We introduce Mesh-TensorFlow, a language for specifying a general class of distributed tensor computations.  Where data-parallelism can be viewed as splitting tensors and operations along the "batch" dimension, in Mesh-TensorFlow, the user can specify any tensor-dimensions to be split across any dimensions of a multi-dimensional mesh of processors.  A Mesh-TensorFlow graph compiles into a SPMD program consisting of parallel operations coupled with collective communication primitives such as Allreduce.  We use Mesh-TensorFlow to implement an efficient data-parallel, model-parallel version of the Transformer \cite{Vaswani17} sequence-to-sequence model.  Using TPU meshes of up to 512 cores, we train Transformer models with up to 5 billion parameters, surpassing state of the art results on WMT'14 English-to-French translation task and the one-billion-word language modeling benchmark.  Mesh-Tensorflow is available at https://github.com/tensorflow/mesh .

\end{abstract}

\section{Introduction}

Batch-splitting (data-parallelism) is the dominant distributed Deep Neural Network (DNN) training strategy, due to its universal applicability and its amenability to Single-Program-Multiple-Data (SPMD) programming.  However, batch-splitting suffers from several major problems when training very large models.  The memory required to store parameters and/or activations and the time necessary to synchronize parameters can make purely-data-parallel algorithms impossible or inefficient.  Different distribution strategies (model-parallelism \cite{Dean:2012:LSD:2999134.2999271}) can solve these issues, but specifying these strategies can be complicated, and the current MIMD implementations generate very large programs which can be difficult to compile and to optimize.

We solve this problem by introducing Mesh-TensorFlow, a language for specifying a general class of distributed tensor computations.  Where data-parallelism can be viewed as splitting tensors and operations along the "batch" dimension, in Mesh-TensorFlow, the user can specify any tensor-dimensions to be split across any dimensions of a multi-dimensional mesh of processors.  A Mesh-TensorFlow graph compiles into a SPMD program consisting of parallel operations coupled with collective communication primitives such as Allreduce.  We use Mesh-TensorFlow to implement an efficient data-parallel, model-parallel version of the Transformer \cite{Vaswani17} sequence-to-sequence model.  Using TPU meshes of up to 512 cores, we train Transformer models with up to 5 billion parameters, surpassing state-of-the-art results on WMT'14 English-to-French translation task and the one-billion-word Language modeling benchmark.



\section{Hardware Assumptions} \label{sec:mesh}
While much work deals with heterogeneous and/or unreliable hardware, we focus on clusters of identical, reliable processors, each with a local memory.  We define a \textbf{mesh} as an n-dimensional array of such processors.  The mesh is only a naming abstraction and does not imply a physical network topology.  As such, different meshes can be defined over the same set of physical processors.  For example, a 512-core TPU cluster with a 16x16x2 toroidal network interconnect could be represented by a 3-dimensional mesh with shape [16, 16, 2], a two-dimensional mesh with shape [32, 16], a one-dimensional mesh with shape [512], etc. The physical network topology \textit{does} affect performance; particularly important is the performance of MPI \textit{Allreduce}, grouped by splitting the mesh by a subset of the dimensions, which can be very efficient \cite{Patarasuk2009} \cite{Jain10y.:optimal} if each such group is physically connected.

\section{Inspiration: Single-Program-Multiple-Data (SPMD) Batch-Splitting} \label{sec:bs}
We first review a commonly-used variant of synchronous data-parallelism where each processor keeps an identical copy of all parameters (Algorithm \ref{alg:bs}).  For each step, the batch of training examples is split into sub-batches, one for each processor.  Each processor computes the forward and backward passes on its sub-batch, resulting in gradients on the model parameters.  These gradients are then summed across all processors and the results broadcast to all processors (MPI-allreduce).  Finally, each processor updates its own copy of the parameters.

\begin{algorithm}[h!]
\caption{Synchronous data-parallelism with replicated parameters.  Each processor maintains a complete copy of all weights $W^{(t)}$. The batch $b^{(t)}$ of training examples for timestep $t$ is partitioned among the set $P$ of processors:  $b^{(t)} = \dot\bigcup_{p\in P} b^{(t)}_p$.   Below is the computation performed on one processor $p \in P$. }
\begin{algorithmic}[1]
\State Compute partial parameter gradients $\nabla Q(W^{(t)}, b^{(t)}_p)$ \Comment{Local computation}
\State $\nabla Q(W^{(t)}, b^{(t)}) = \sum_{p' \in P} \nabla Q(W^{(t)}, b^{(t)}_{p'})$ \Comment{$Allreduce$}
\State $W^{(t+1)} = Update(W^{(t)}, \nabla Q(W^{(t)}, b^{(t)})$ \Comment{Local computation}
\end{algorithmic}
\label{alg:bs}
\end{algorithm}

This algorithm is typically implemented using Single-Program-Multiple-Data (SPMD) programming, with every processor running the same program of local operations and MPI-allreduce primitives.

One way to see this algorithm is that every tensor and every operation in the computation is either split across all processors (if it has a "batch" dimension), or fully replicated across all processors (if it does not have a "batch" dimension). Operations which reduce out the "batch" dimension require an additional MPI-allreduce to produce the correct result.  We can describe this as splitting the computation across the "batch" dimension.  Mesh-TensorFlow generalizes this idea to splitting computations across arbitrary dimensions.

\section{Mesh-TensorFlow: Beyond Batch Splitting}
Mesh-Tensorflow generalizes from the batch-splitting algorithm described in section \ref{sec:bs} to allow for splitting across different Tensor dimensions.  The similarities are as follows:

\begin{itemize}

\item Each tensor in the computation is represented by one (not-necessarily-distinct) slice of the tensor on each processor.

\item Each operation in the computation is implemented as one operation on each processor.  Most operations require no communication, with each processor producing its slice of the output from its slices of the inputs.  Some operations additionally require collective communication primitives such as MPI-allreduce.

\item Based on the above, the computation can be implemented as a SPMD program.

\end{itemize}

The new elements in Mesh-TensorFlow are as follows:

\begin{itemize}

\item Tensors have named dimensions.  This allows for the idea of a logical dimension (like "batch") which will be split in the same way for different tensors and operations.  It is illegal for a tensor to have two identically-named dimensions. 

\item Rather than an unstructured set of processors, Mesh-Tensorflow allows for an n-dimensional mesh of processors (section \ref{sec:mesh}).  The mesh also has named dimensions.

\item A global "computation layout" is a partial map from tensor-dimension to mesh-dimension specifying which tensor-dimensions are split across which dimensions of the processor-mesh.  For example, batch-splitting (data-parallelism) would be expressed by using a one-dimensional mesh with dimension \texttt{"all\_processors"} and using the computation layout \texttt{[("batch", "all\_processors")]}.  This means that all tensors with a \texttt{"batch"} dimension are split along that dimension across all processors, while all other tensors are fully replicated.  

\end{itemize}


\section{Tensor Representations}
A tensor is represented as one slice of the tensor per processor.  The \textbf{layout} of a tensor is an injective partial map from the tensor's dimensions to dimensions of the mesh, and is computed as the restriction of the global computation layout to that tensor's dimensions.  It is illegal for two dimensions of the same tensor to map to the same mesh dimension.  If a tensor's layout is empty, it is fully replicated on each processor.  For every (tensor-dimension, mesh-dimension) pair in the tensor's layout, the slice on a processor is restricted along that tensor-dimension to a stripe corresponding to that processor's coordinate along that mesh-dimension.   The current implementation of Mesh-TensorFlow requires the size of the tensor-dimension to be evenly divisible by the size of the mesh-dimension.

\section{Operation Implementation}
Each operation is implemented by parallel computation on every processor, and sometimes collective communication.  We describe the implementations of some important operations here:

\paragraph{Component-wise Operations} Mesh-TensorFlow supports component-wise operations where the shapes (and hence the layouts) of the input and output tensors are identical.  These are trivially implemented by parallel operations on each processor to compute that processor's slice of the output from that processor's slice(s) of the input(s).

\paragraph{Reduction (reduce\_sum(), reduce\_max(), etc.)}  Mesh-TensorFlow supports reductions where the output dimensions are a subset of the input dimensions.  These can be implemented by local reductions of each slice, followed by MPI-allreduce across any mesh dimensions corresponding to reduced-out Tensor dimensions.  The allreduce operation is necessary because the local reduction only sums across a subset of the split tensor-dimension.  Bandwidth-efficient implementations of allreduce exist when the processors for each group are connected in any type of tree. \cite{Patarasuk2009} \cite{Jain10y.:optimal}

\paragraph{Einstein Summation (matrix multiplication, etc.)}  Einstein-summation (einsum) notation (as defined in numpy, TensorFlow, etc.) is a way of expressing a class of operations including (batch) matrix multiplication, reductions and broadcasts, where the operation is defined by the names of the dimensions of the input and output tensors.  Mesh-TensorFlow's use of named dimensions makes using einsum particularly convenient.  Einsum can be defined as broadcasting all inputs to a shape consisting the union of all their dimensions, multiplying them component-wise, then reducing out all dimensions not in the specified output shape.   Einsum is implemented by parallel einsum operations on each processor of that processor's input slices, followed by MPI-allreduce across any mesh dimensions corresponding to reduced-out Tensor dimensions.

\subsection{Reshape}  While reshape is simple in the non-distributed case, Mesh-TensorFlow reshape can require network communication, since the layout of the output tensor may differ from that of the input tensor.  Even keeping the same dimension sizes, changing the dimension names (and hence the layout) can result in several different communication patterns:  If a dimension is split in the input but not in the output, the implementation involves MPI-allgather communication across the corresponding mesh-dimension.  If a dimension is split in the output but not in the input, the implementation involves no communication, just slicing on each processor.  MPI-alltoall is used in the case where different dimensions in the input and the output are split across the same mesh dimension, as might be the case when switching between data-parallelism and model-parallelism for different layers of the same model, as in \cite{Shazeer17}.

\section{Mesh-TensorFlow syntax}
The Mesh-TensorFlow language is nearly identical to TensorFlow \cite{tensorflow2015-whitepaper}, with the familiar notions of graphs, tensors, operations, variables, devices (called meshes), and automatic gradient computation.  The principal difference is that in Mesh-TensorFlow, tensor-dimensions have a name as well as a size.  The shape of each tensor is a statically-known tuple of such dimensions.   Shapes are inferred automatically when possible, as they are in TensorFlow.  Binary component-wise operations like addition employ implicit broadcasting in the case where the shape of one operand is a subset of the shape of the other.

The initial implementation of Mesh-TensorFlow is a Python library.  The user builds a Mesh-TensorFlow graph in python, which the library "lowers" to generate part of a TensorFlow graph.  As of the writing of this paper, implementations exist for generating SPMD TensorFlow code for TPUs, or MIMD code (using device placement) for multi-CPU/GPU configurations.



\section{Example: Two Fully-Connected Layers} \label{example}
We consider a simple example of two fully-connected layers in the middle of a neural network.  The input layer $x$ and the output layer $y$ each have $d_{io}$ units, and the hidden layer $h$ has $d_h$ units.  The hidden layer also has a bias and $Relu$ activation.

\begin{equation}
    y = Relu(xw + bias)v
\end{equation}

This Mesh-TensorFlow code fragment runs these layers on a batch $x$ of $batch\_size=b$ inputs.

\begin{verbatim}
  ...
  batch = mtf.Dimension("batch", b)
  io = mtf.Dimension("io", d_io)
  hidden = mtf.Dimension("hidden", d_h)
  # x.shape == [batch, io]
  w = mtf.get_variable("w", shape=[io, hidden])
  bias = mtf.get_variable("bias", shape=[hidden])
  v = mtf.get_variable("v", shape=[hidden, io])
  h = mtf.relu(mtf.einsum(x, w, output_shape=[batch, hidden]) + bias)
  y = mtf.einsum(h, v, output_shape=[batch, io])
  ...
\end{verbatim}

The code above defines only the mathematical model.  We now discuss several different computation layouts.  Each will produce identical results, but will have different performance characteristics. We also provide illustrations of the layouts in Appendix \ref{sec:illustrations}.

\subsection{Data-Parallel Layout}
To train the above model in data-parallel mode on a mesh of $n$ processors, we would define:
\begin{verbatim}
  mesh_shape = [("all", n)]
  computation_layout = [("batch", "all")]
\end{verbatim}
When the Mesh-TensorFlow graph is compiled with this layout, the parameter tensors $w$, $v$, and $bias$ are replicated on all processors, but the activation matrices $x$, $h$, $y$, etc. are split across the batch dimension.  For example, each processor keeps a slice of $x$ with shape $[\frac{b}{n}, d_{io}]$.

There is no inter-processor communication in the forward pass. However, the gradient computations for the parameters are \texttt{mtf.einsum} operations which reduce out the \texttt{batch} dimension, and hence produce $Allreduce$ operations when they are compiled.  The number of values allreduced per processor is equal to the number of parameters, approximately $2d_{io}d_h$.

\subsection{Model-Parallel Layout} \label{exmp}
Rather than splitting the batch, we can split the units in the hidden layer:
\begin{verbatim}
  mesh_shape = [("all", n)]
  computation_layout = [("hidden", "all")]
\end{verbatim}
When the Mesh-TensorFlow graph is compiled with this layout, the input and output layers $x$, and $y$ are replicated on all processors, but the hidden activations $h$ and the parameter tensors $w$, $v$ and $bias$ are all split across the \texttt{hidden} dimension.   For example, each processor keeps a slice of $w$ with shape $[d_{io}, \frac{d_h}{n}]$ and a slice of $v$ with shape $[\frac{d_h}{n}, d_{io}]$.

When computing $y$, the split hidden dimension is reduced out.  Consequently, the results of that computation get allreduced across all processors.  A similar allreduce happens in computing the gradients on $x$.  In all, the number of values allreduced per processor is $2bd_{io}$.

\subsection{Data-Parallel, Model-Parallel Layouts}
On a two-dimensional mesh of $r \times c$ processors, we can employ both data-parallelism and model-parallelism:
\begin{verbatim}
  mesh_shape = [("rows", r), ("cols", c)]
  computation_layout = [("batch", "rows"), ("hidden", "cols")]
\end{verbatim}

In this layout, each row of processors handles a fraction of the batch, while each column of processors handles a fraction of the hidden units.  Each processor keeps a slice of x with shape $[\frac{b}{r}, d_{io}]$, with processors in the same row having identical slices.  The hidden activation tensor $h$ is tiled in two dimensions, with each processor keeping a slice with shape $[\frac{b}{r}, \frac{d_h}{c}]$.

This layout causes partitioned-allreduce operations in several places.  For example, in computing $y$, we reduce out the \texttt{hidden} dimension, which is split over the \texttt{cols} dimension of the mesh, so the results of the operation need to be summed up by processor-column, as opposed to over the entire mesh.  In all, the number of values allreduced per processor is $\frac{2bd_{io}}{r} + \frac{2d_{io}d_h}{c}$

If we have a three-dimensional mesh of processors, we can even split the computation in three dimensions:
\begin{verbatim}
  mesh_shape = [("rows", r), ("cols", c), ("planes", p)]
  computation_layout = [
    ("batch", "rows"), ("hidden", "cols"), ("io", "planes"])
\end{verbatim}
In this case, every matrix in the computation is tiled across two mesh dimensions and replicated in the third, and every \texttt{einsum} requires an allreduce across one mesh dimension.

\subsection{Inefficient Layouts}
For a computation layout to be efficient, all expensive operations need to be split (as opposed to replicated) across all mesh dimensions.  For example, the empty layout below produces correct results, but since it replicates all computation on every processor, it saves no time or memory.  A general rule is that any expensive \texttt{einsum} operation should have one input dimension that is split across each batch dimension.
\begin{verbatim}
  mesh_shape = [("all", n)]
  computation_layout = []
\end{verbatim}

\subsection{Illegal Layouts}
The computation layout below is illegal, because it causes the tensor $h$ to have two dimensions which are split across the same dimension of the mesh.
\begin{verbatim}
  mesh_shape = [("all", n)]
  computation_layout = [("batch", "all"), ("hidden", "all")]
\end{verbatim}

\subsection{Performance Comparison}

\def \DXY {d_{io}}
\def \COMP {b \DXY d_h}

\begin{table}[h!] \label{tab:performance}
\scalebox{0.9}{
\begin{tabular}{ |c|c|c|c|c|}
\hline
 Layout & Comp. Time & Comm. Time & $\frac{communication}{computation}$ & Memory/Processor \\
 \hline\hline

\rule{0pt}{3ex} \texttt{[]} & $\COMP$ 
   & $0$
   & $0$ 
   &  $b\DXY + bd_h + \DXY d_h $ \\
\hline
\rule{0pt}{3ex} \texttt{[("batch", "all")]} & $\frac{\COMP}{n} $ 
   & $\DXY d_h$
   & $\frac{n}{b}$ 
   &  $\frac{b}{n} \DXY + \frac{b}{n} d_h + \DXY d_h $ \\
\hline
\rule{0pt}{3ex} \texttt{[("hidden", "all")]}  & $\frac{\COMP}{n} $ 
   & $\DXY b$
   & $\frac{n}{d_h}$ 
   &  $b \DXY + b \frac{d_h}{n} + \DXY \frac{d_h}{n} $ \\
\hline
\rule{0pt}{3ex}  \makecell{\texttt{[("batch", "rows"),}\\ \texttt{ ("hidden", "cols")]}} & $\frac{\COMP}{rc} $
   & $\DXY (\frac{b}{r} + \frac{d_h}{c} )$
   & $\frac{c}{d_h} + \frac{r}{b}$ 
   & $ \frac{b}{r} \DXY + \frac{b}{r} \frac{d_h}{c} + \DXY \frac{d_h}{c} $ \\
\hline
\rule{0pt}{3ex}  \makecell{\texttt{[("batch", "rows"),} \\ \texttt{ ("hidden", "cols"),} \\ \texttt{ ("io", "planes")]}}  & $\frac{\COMP}{rcp} $
   & \scalebox{0.7}{$\frac{b}{r} \frac{\DXY}{p} + \frac{b}{r} \frac{d_h}{c} + \frac{\DXY}{p} \frac{d_h}{c}$}
   & \scalebox{0.9}{$\frac{c}{d_h} + \frac{p}{\DXY} + \frac{r}{b} $}
   & $ \frac{b}{r} \frac{\DXY}{p} + \frac{b}{r} \frac{d_h}{c} + \frac{\DXY}{p} \frac{d_h}{c}$ \\
\hline
\end{tabular}
}
\caption{Computation, communication and memory costs for different layouts of the computation in Algorithm 1.  Constant factors and lower-order terms are dropped.}
\label{tab:performance}
\end{table}

Table \ref{tab:performance} shows the computational costs associated with our example computation layouts.  The computation time is dominated by that of \texttt{einsum} operations.  The communication time comes from the \textit{Allreduce} operations, which are necessary whenever the inner dimension of einsum is split.  Assuming that the mesh has physical links between all pairs of logically adjacent processors, each \textit{Allreduce} operations can be done in time proportional to the size of one slice divided by the per-link network bandwidth \cite{Jain10y.:optimal}.  

The network-boundedness of the computation is proportional to the value shown in the table column marked $\frac{communication}{computation}$, with the constant of proportionality depending on the ratio of communication and computation speeds on the given hardware.   In the data-parallel layout, the value is $\frac{n}{b}$, the inverse of the per-processor batch size.  Performance suffers if the per-processor batch is too small.  In the model-parallel layout, the value is $\frac{n}{d_h}$, the inverse of the number of hidden units per processor.  Performance suffers if the hidden layer is sliced too finely.  For good performance, batch size is irrelevant, but we need the hidden layer to get larger as we increase the number of processors.  In the first data-parallel, model-parallel layout, the value is $\frac{c}{d_h} + \frac{r}{b}$.  In this layout, we can quadratically increase the number of processors while only linearly increasing the batch size and hidden layer sizes necessary to maintain good efficiency.  The final layout lets us cubically increase the number of processors in a 3-dimensional mesh, while only linearly increasing the batch size and the layer sizes.

\section{Model-Parallel "Transformer"} \label{sec:transformer}

We implemented a model-parallel layout of the Transformer attention-based sequence-to-sequence model described in \cite{Vaswani17}.  The complete implementation is available in the \textit{tensor2tensor} library on \textit{github}.  The layout is given by:
\begin{verbatim}
  mesh_shape = [("all", n)]
  computation_layout = [
    ("vocab", "all"), ("d_ff", "all"), ("heads", "all")]
\end{verbatim}
That is, the dimensions representing the vocabulary size, the size of the feed-forward hidden layer, and the number of attention heads are each split across all processors.  This layout works because every expensive operation in the model has exactly one of these dimensions, and no tensor in the model has more than one.  Similarly to the model-parallel layout for our example network (Section \ref{exmp}), network-boundedness and memory usage per processor remain constant if we scale all of these dimensions proportionally to the number of processors.   We did just this, training transformer models with ever larger hidden layers and numbers of attention heads on ever larger TPU clusters (we did not increase the vocabulary size).  As expected, we saw very similar performance characteristics between the models.  This scaling turns out to be highly beneficial to model quality (Section \ref{sec:results}).

To use even more processors, we combined this model-parallelism with data parallelism, splitting the batch across one dimension of a 2-dimensional TPU mesh and the dimensions described above across the other dimension of the mesh:
\begin{verbatim}
  mesh_shape = [("rows", r), ("cols", c")]
  computation_layout = [("batch", "rows"), ("vocab", "cols"),
                        ("d_ff", "cols"), ("heads", "cols")]
\end{verbatim}
This layout maintains constant performance if the batch size is scaled proportionally to r and the mentioned model dimensions are scaled proportionally to c.  Using this layout, we trained Transformer models with feed-forward hidden dimensions up to 262144 and up to 256 attention heads on 2-dimensional TPUv2 meshes of up to 16x32=512 cores, maintaining computational efficiency of over 50\% (6 PFLOP/s out of a maximum 11.5 PFLOP/s) on the largest models.

\subsection{Experiments and Results} \label{sec:results}
To examine the benefit of scaling the Transformer model in the manner suggested by the previous section, we trained such models on machine translation and language modeling tasks.  Results are given in Tables \ref{tab:lm} and \ref{tab:translation}.   

For the billion-word language modeling benchmark, we trained the models for 10 epochs.  The largest model (4.9B parameters) took 13 hours to train on a 512-core TPUv2 cluster.  Batch size for all models was 256 sequences of 256 tokens each (each sequence was the concatenation of multiple training sentences).  The batch was split along the mesh dimension of size 16 and the model dimensions were split along the mesh dimension of size 32.  Per-word dev-perplexity for the largest model was 24.0, but dropped to 23.5 when the model was evaluated with the logits multiplied by 0.9 (likely due to overfitting).  This represents the best published result on this dataset.  As expected, perplexity was lower for larger models.  We have included random samples from these models in Appendix \ref{samples}.   On the \texttt{languagemodel\_wiki\_noref\_v128k\_l1k} dataset from the Tensor2Tensor library\footnote{No published results exist for this dataset.}, consisting of over 5 billion tokens of text from Wikipedia, perplexity continued to improve significantly with a model size of 5 billion parameters.  

On the WMT14 En-Fr translation tasks (\ref{tab:translation}), we trained the models for 3 epochs.   The largest model (2.9B parameters) was trained for 22 hours on a 128-core TPUv2 cluster.  Quality improved with model size, with the largest model achieved BLEU score 43.9 (evaluated using \texttt{sacrebleu}), the best published result to date.  For the WMT14 En-De dataset, gains from model size were smaller, presumably due to the small size of the training data.

Additional details about the configurations for these experiments are available as part of the \texttt{tensor2tensor} library on github.


\begin{table}[h!]
\caption{Transformer-Decoder Language Models:   $d_{model}=1024$, $d_k = d_v = 256$}
\label{tab:lm}
\begin{center}
\vspace{-2mm}
\scalebox{0.8}{
\begin{tabular}{ccc|cc}
\hline\rule{0pt}{2.0ex}
$d_ff$ & $heads$ & Parameters & Billion-Word Benchmark & Wikipedia \\
& & (Billions) & Word-Perplexity & Subword-Perplexity \\
\hline
4096 & 4 & 0.14 & 35.0 & 8.74 \\
8192 & 8 & 0.22 & 31.7 & 8.03 \\
16384 & 16 & 0.37 & 28.9 & 7.44 \\
32768 & 32 & 0.67 & 26.8 & 6.99 \\ 
65516 & 64 & 1.28 & 25.1 & 6.55 \\ 
131072 & 128 & 2.48 & 24.1 & 6.24 \\
262144 & 256 & 4.90 & 24.0(\textbf{23.5}) & \textbf{6.01} \\
\hline
\multicolumn{2}{l}{Prev Best DNN \cite{Shazeer17}} & ~6.5 & 28.0 \\
\multicolumn{2}{l}{Best DNN Ensemble \cite{Jozefowicz16}} &  & 26.1 \\
\multicolumn{2}{l}{Best Ensemble (different methods)\cite{Jozefowicz16}} & $>100$ & 23.7 \\
\hline
\end{tabular}
}
\end{center}
\end{table}

\begin{table}[h!]
\caption{Transformer Machine-Translation Results.  $d_{model}=1024$, $d_k = d_v = 128$ }
\label{tab:translation}
\begin{center}
\vspace{-2mm}
\scalebox{0.8}{
\begin{tabular}{cccc|ccc}
\hline\rule{0pt}{2.0ex}
$d_ff$ & $heads$ & $d_k, d_v$ & Parameters & WMT14 EN-DE & WMT14 EN-FR \\
& & & (Billions) &BLEU & BLEU \\
\hline
2048 & 4 & 128 & 0.15 & 25.5 & 41.8 \\
4096 & 8 & 128 & 0.24 & 26.5 & 42.5 \\
8192 & 16 & 128 & 0.42 & 27.1 & 43.3 \\
16384 & 32 & 128 & 0.77 & 27.5 &  43.5 \\
32768 & 64 & 128 & 1.48 & 27.5 &  43.8 \\ 
65536 & 128 & 128 & 2.89 & 26.7 &  \textbf{43.9} \\ 
\hline
4096 & 16 & 64 & 0.21 & \textbf{28.4} & 41.8 & \cite{Vaswani17} \\ 
\hline
\end{tabular}
}
\end{center}
\end{table}

\section{Related Work}
A large part of deep learning computations is a series of matrix multiplications and tensor contractions (\emph{Einsum}s). Distributed matrix multiplication is a well-studied problem in high performance computing. Efficient algorithms partition the computational space, instead of partitioning work by the output matrix/tensor (\emph{owners compute}), to minimize communication. This technique is sometimes called \emph{iteration space tiling}~\cite{Wolfe:1989:MIS:76263.76337}, \emph{replication}~\cite{SD_EUROPAR_2011}, or \emph{task parallelism}~\cite{Calvin:2015:STA:2833179.2833186}. 
Mesh-TensorFlow can express a wide range of uniform partitionings of the iteration space 
and therefore can adopt many best known mappings, 
e.g., 3D~\cite{Aggarwal:1990:CCP:77831.77836, berntsen1989communication} and 2.5D~\cite{SD_EUROPAR_2011} algorithms for square matrices, CARMA~\cite{demmel2013communication} for rectangular matrices, 1.5D~\cite{SpDM3} algorithm for matrices with different sparsities, best tile sizes for direct convolutions~\cite{demmel2018communication}, etc., although sometimes with higher memory requirements. 
Furthermore, in most existing work,
when multiple multiplications are composed together, the user has to specify the data layout for each matrix separately~\cite{pmlr-v84-koanantakool18a}. Mesh-TensorFlow lets the user name the dimension to split, simplifying the process and allowing for much easier mapping explorations. Feature-wise, Mesh-TensorFlow shares many similarities with the Cyclops Tensor Framework~\cite{solomonik2014massively}, a distributed tensor contraction library originally developed for quantum chemistry applications, which also supports replication and arbitrary mappings.

In the context of deep learning, partitioning the iteration space, e.g., interpolating between data and model parallelism, is relatively new. Gholami et al.~\cite{gholami2017integrated} analytically showed that using both data and model parallelism at the same time can be more beneficial than using just one of them. Building on top of 1.5D matrix multiplication algorithms, their algorithm can support \emph{replication} and arbitrary processor grid shapes. However, they only explored the parallelization of AlexNet~\cite{krizhevsky2012imagenet} and they have not implemented the algorithm. Jia et al.~\cite{jia2018exploring, jia2018beyond} implemented a framework that uses cost modeling to pick the best parallelization strategy, including how to partition work for each operation. Their \emph{parallelizable dimensions} are defined as the set of all divisible dimensions in the output tensor (\emph{owners compute}), and therefore their mapping can be suboptimal in terms of communication. We expand on this in Appendix \ref{sec:costs}. 


\section{Future Work}
The Mesh-TensorFlow library is available at \texttt{https://github.com/tensorflow/mesh} and is under active development.  Some potential areas for development are:
\begin{itemize}
    \item Automated search for optimal computation layout.
    \item Implementations of different models and operations.  For example, convolutions on spatially-partitioned tensors will require the communication of "halo" regions, as described in \cite{Jin18}.
    \item Implementation of SPMD programming on CPU/GPU clusters.
\end{itemize}

\section{Conclusion}
In this paper, we introduce the Mesh-TensorFlow language, facilitating a broad class of SPMD distributed tensor computations.  Applying Mesh-TensorFlow to the Transformer model, we are able to train models with 5 billion parameters on up to 512-core clusters, establishing new state-of-the-art results for WMT14 En-Fr translation task and the One Billion Word language modeling benchmark.

\printbibliography
\newpage
\appendix
\input{appendix}

\end{document}

%% file: appendix
\section{Illustrations for the Two Fully-Connected Layers Example} \label{sec:illustrations}
\begin{figure}[h]
    \centering
    \includegraphics[page=4,width=8cm,trim={0cm 0cm 5cm 2.7cm},clip]{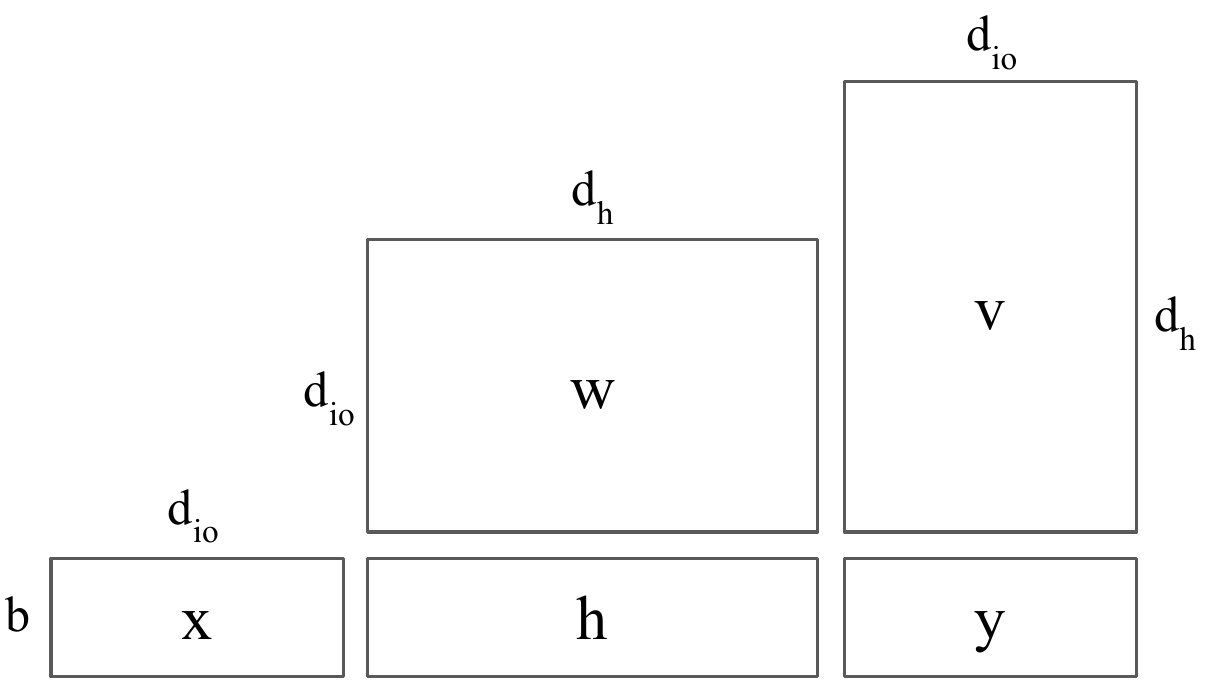}
    \caption{The overall computation of the Two Fully-Connected Layers example. First, $x$ is multiplied with $w$. The temporary result $xw$ is summed with $bias$, applied component-wise $Relu$, and then stored in $h$. Lastly, we multiply $h$ with $v$ to get $y$. The component-wise operations and the intermediate matrix $xw$ are not shown in the figure.}
    \label{fig:ex:overall}
\end{figure}

This section provides the illustrations of the four layouts mentioned in the Two Fully-Connected Layers example in Section \ref{example}. The overall computation is shown in Figure~\ref{fig:ex:overall}. We draw a matrix multiplication $C=AB$ by putting $A$ to the left of $C$ and putting $B$ above $C$. For each matrix, we put its name inside, its number of rows on the left or right side, and its number of columns above or below it. We omit the numbers of rows or columns that can be implied from adjacent matrices, i.e., knowing that the multiplication dimensions must match.

\begin{figure}[t]
    \centering
    \includegraphics[page=5,width=9cm,trim={0cm 0cm 4.3cm 2.9cm},clip]{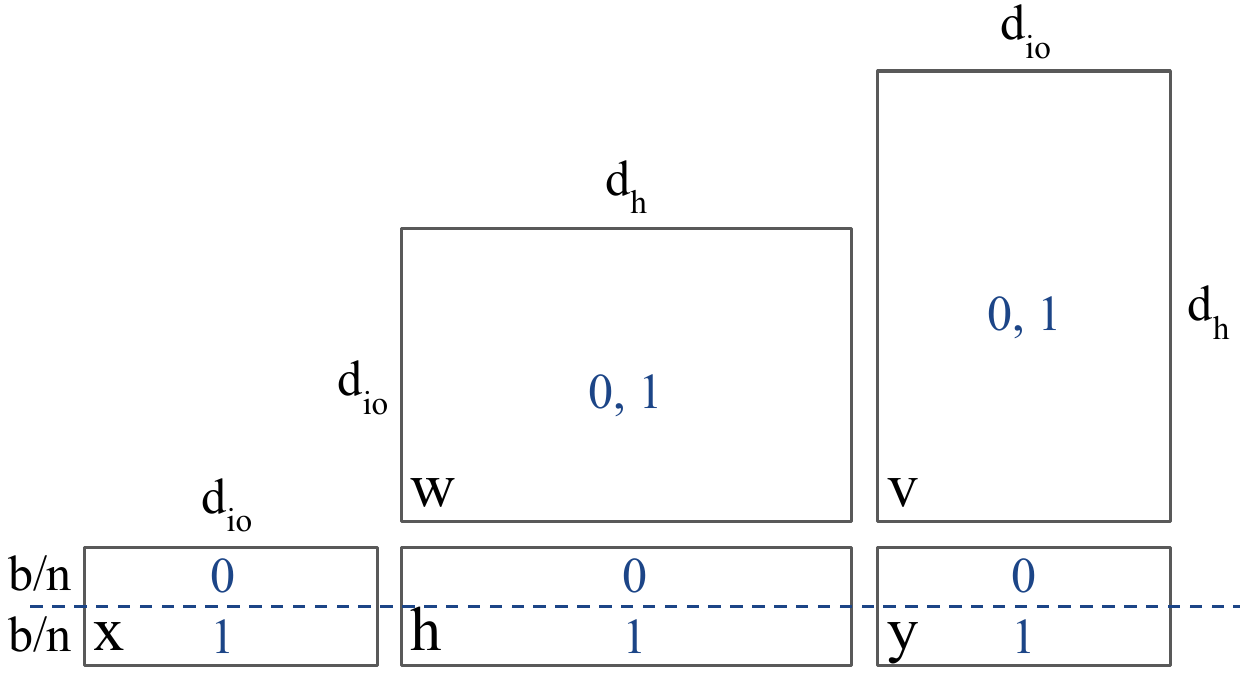}
    \caption{The data-parallel layout for the Two Fully-Connected Layers example, with $n=2$ processors $\in \{0, 1\}$. Blue numbers on matrices indicate the ranks of the processors the matrix slices reside on. The $batch$ dimension is split among all processors. $w$ and $v$ are fully replicated.}
    \label{fig:ex:data-parallel}
\end{figure}

\begin{figure}[t]
    \centering
    \includegraphics[page=6,width=9cm,trim={0cm 0.5cm 3.7cm 1.7cm},clip]{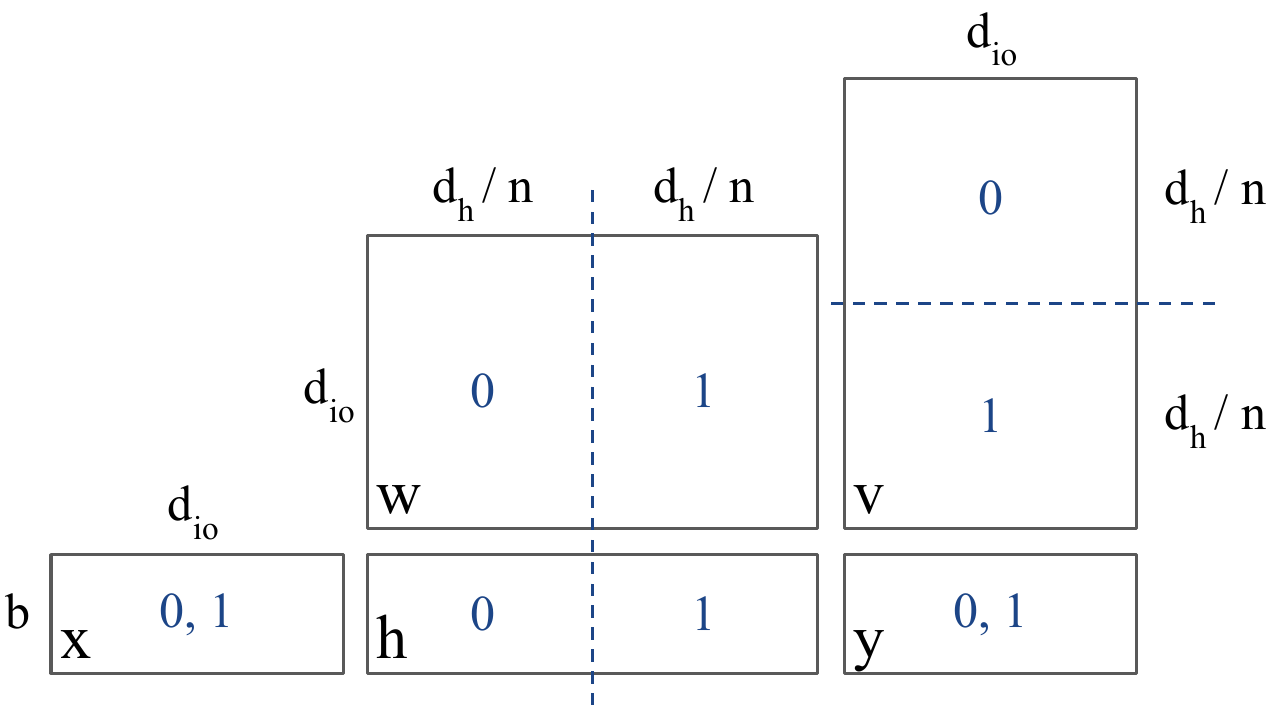}
    \caption{The model-parallel layout for the Two Fully-Connected Layers example, with $n=2$ processors $\in \{0, 1\}$. Blue numbers on matrices indicate the ranks of the processors the matrix slices reside on. The $hidden$ dimension is split among all processors. $x$ and $y$ are fully replicated.}
    \label{fig:ex:model-parallel}
\end{figure}

Figure~\ref{fig:ex:data-parallel} presents the purely data-parallel layout with $n=2$ processors. We number the processors 0 and 1, respectively. The ranks of the processors that store each matrix part are written in blue on the matrix part. The matrix names are moved to the bottom-left corners. The whole $w$ and $v$ are labeled with both 0 and 1 because both of them are fully replicated between the two processors.
The purely model-parallel layout are drawn similarly in Figure~\ref{fig:ex:model-parallel}.

\begin{figure}
    \centering
    \includegraphics[page=7,width=10cm,trim={0cm 0cm 3.3cm 2.2cm},clip]{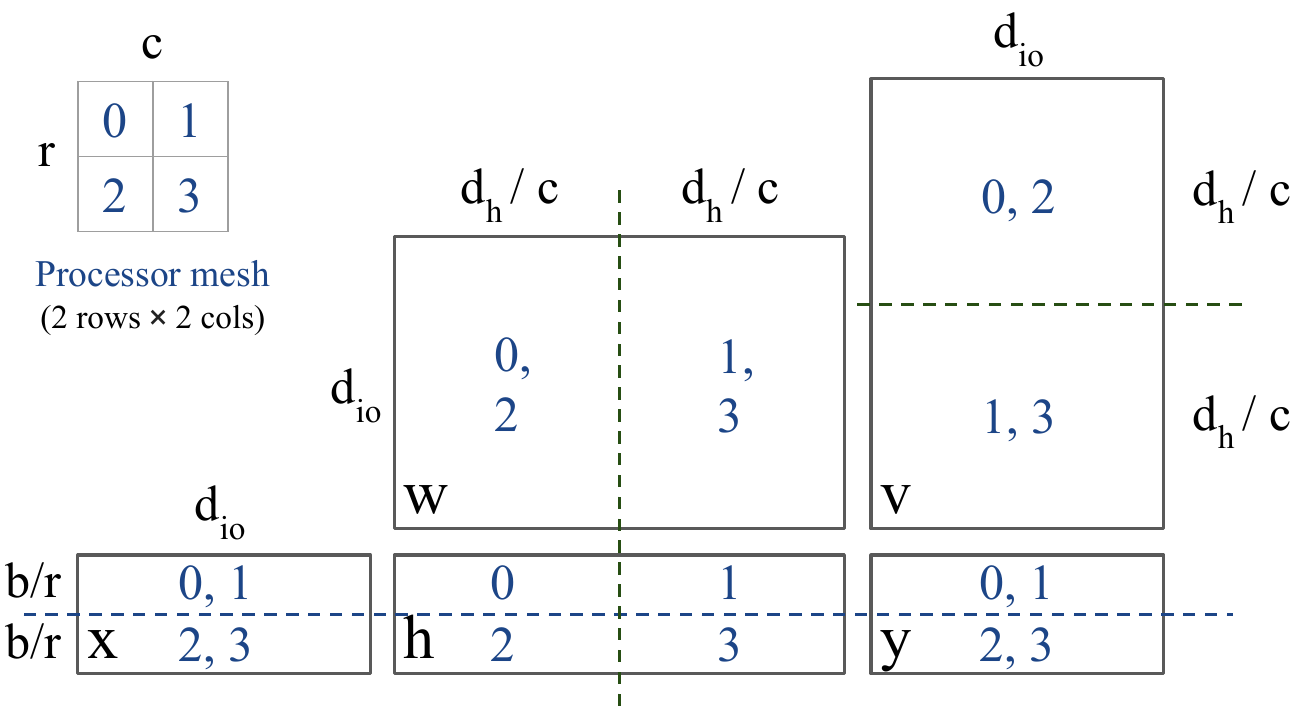}
    \caption{The mixed data-and-model-parallel layout for the Two Fully-Connected Layers example. There are 4 processors, arranged into a 2-by-2 mesh. Each processor is assigned a serialized rank which is used to label matrix slices that it owns.}
    \label{fig:ex:mixed-parallel-2d-mesh}
\end{figure}

\begin{figure}[!t]
    \centering
    \includegraphics[page=8,width=10cm]{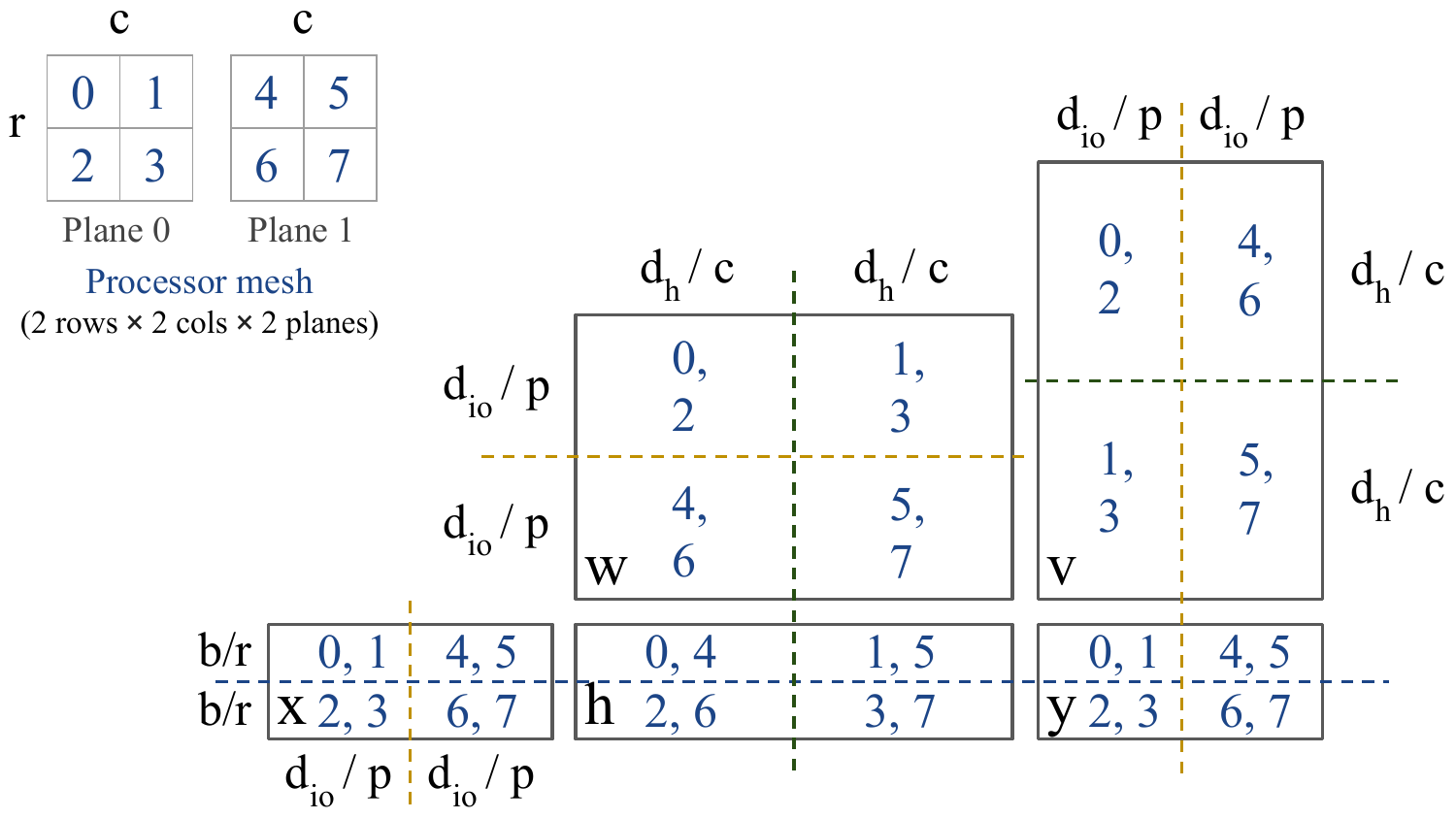}
    \caption{The mixed data-and-model-parallel layout for the Two Fully-Connected Layers example. There are 8 processors, arranged into a 2-by-2-by-2 mesh. Each processor is assigned a serialized rank which is used to label matrix slices that it owns.}
    \label{fig:ex:mixed-parallel-3d-mesh}
\end{figure}

Figures~\ref{fig:ex:mixed-parallel-2d-mesh} and~\ref{fig:ex:mixed-parallel-3d-mesh} show the mixed data-and-model-parallel layout with a 2-by-2 and a 2-by-2-by-2 processor meshes, respectively. We give each processor a serialized rank as shown in the figure, and use the serialized rank to label matrix slices. 

\FloatBarrier

\section{Per-operation Parallelizations and Communication Costs}\label{sec:costs}
\label{section:iteration_space}
Communication is much more expensive than computation and is usually the bottleneck in a parallel program, especially in distributed setting.
The section shows how the more common \emph{owner-compute} parallelization strategies can be communication-suboptimal.
We start with a simplified overview of the parallelization schemes used in distributed matrix multiplication and tensor contractions (\emph{Einsum}s), focusing on their communication bandwidth costs. (See~\cite{irony2004communication, ballard2011minimizing, hbl} for rigorous communication lower bounds analyses.) We only discuss distributed matrix multiplication here since the concept is trivially generalizable to its tensor counterpart.

\begin{figure}[h]
    \centering
    \includegraphics[page=1,width=9cm,trim={0.5cm 1.5cm 1.5cm 1.75cm},clip]{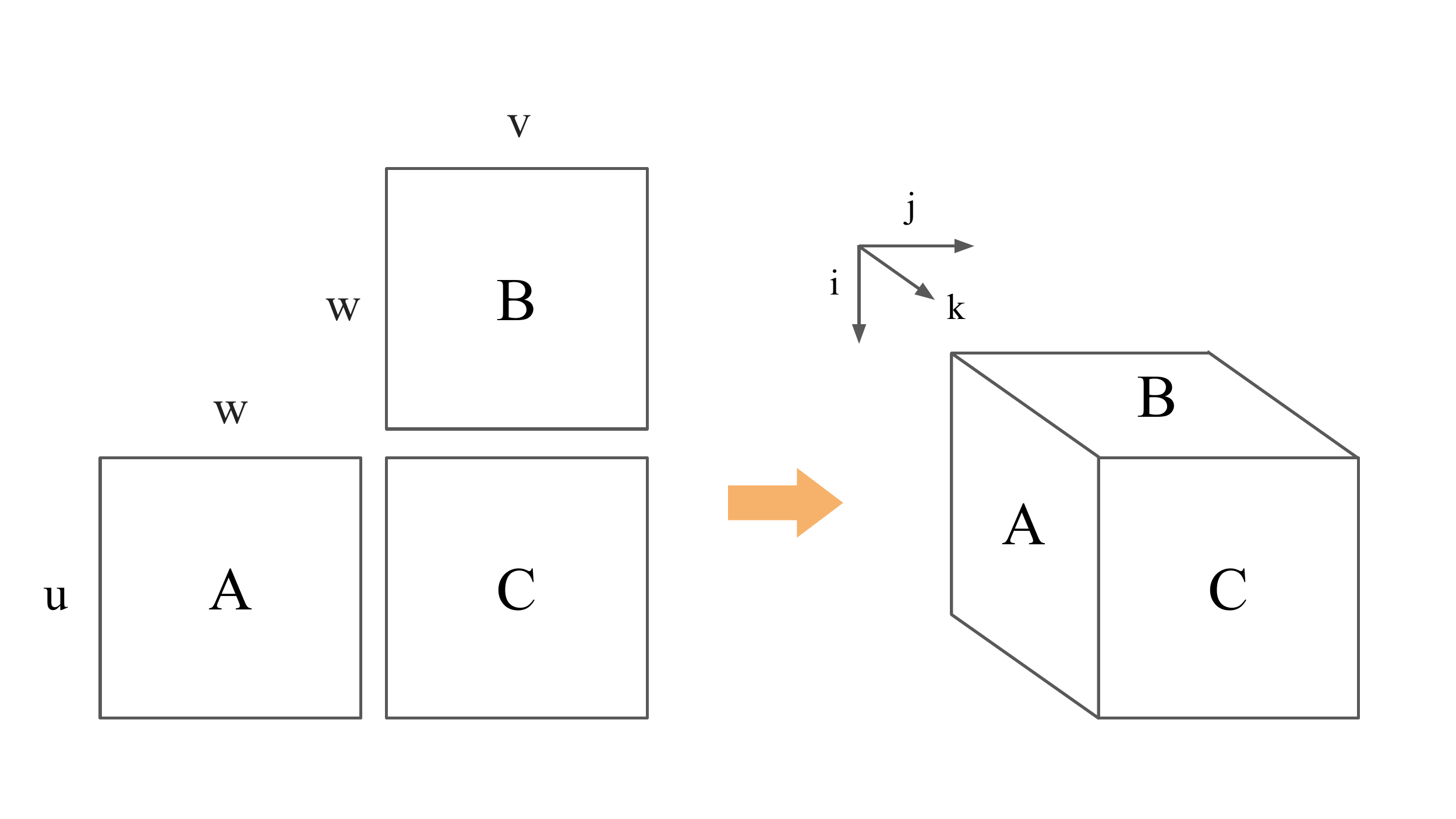}
    \caption{The iteration space of the matrix multiplication problem $C=AB$. Each voxel $(i,j,k)$ represents the computation $c_{ij}\ {+}{=}\ a_{ik}b_{kj}$.}
    \label{fig:iteration_space}
\end{figure}

\begin{figure}[t]
    \centering
    \includegraphics[page=3,height=4cm,trim={0.5cm 0cm 11.8cm 0.5cm},clip]{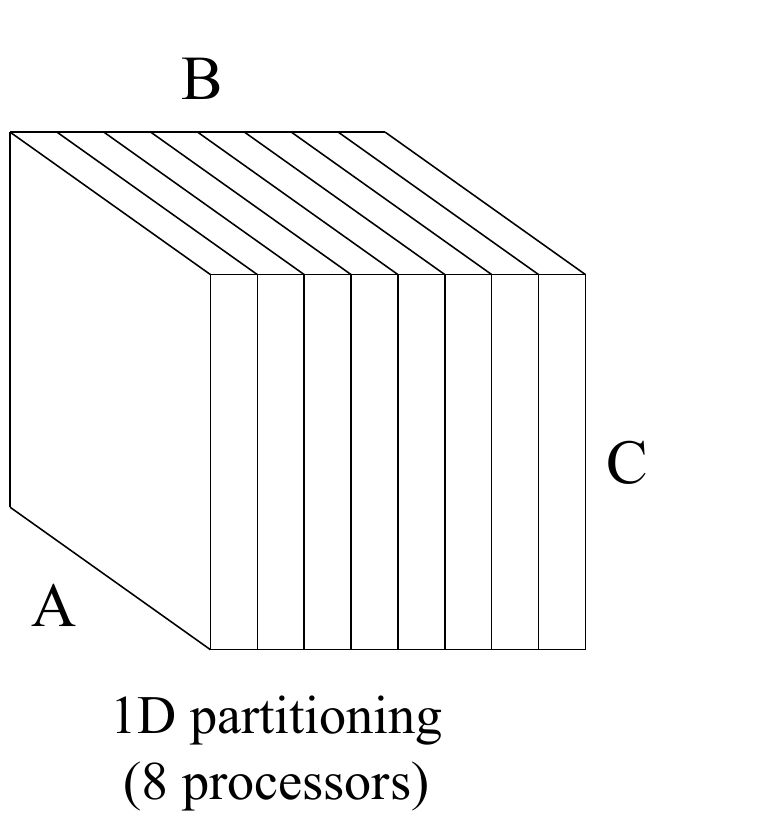}
    \includegraphics[page=2,height=4cm,trim={0.5cm 0cm 1.5cm 0.5cm},clip]{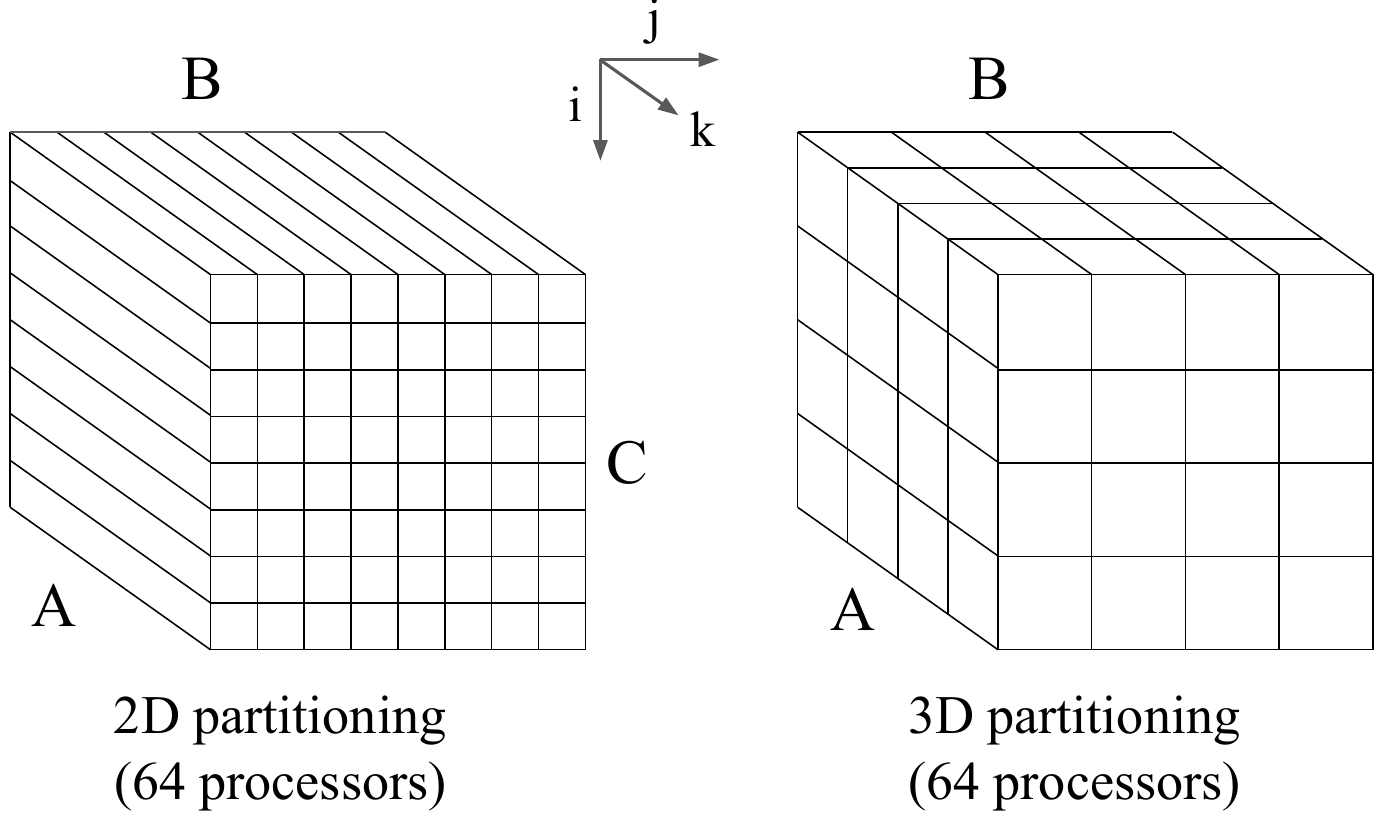}
    \caption{Example partitionings of the iteration space for the matrix multiplication problem. The names 1D, 2D, and 3D comes from the number of dimensions that are split.}
    \label{fig:1d2d3d}
\end{figure}

\paragraph{Iteration Space.} The iteration space is the set of all index tuples required to compute a problem. For example, the matrix multiplication problem $C=AB$ computes $c_{ij} = \sum_k a_{ik}b_{kj}$. Its iteration space consists of all possible tuples $(i, j, k) \in \mathbb{Z}^3, 0 \le i < u, 0 \le j < v, 0 \le k < w$, where $C$ is $u$-by-$v$, $A$ is $u$-by-$w$, and $B$ is $w$-by-$v$, as shown in Figure~\ref{fig:iteration_space}.
\paragraph{Parallelization.} Let $n$ be the number of processors. Parallelization corresponds to partitioning the set of voxels into $n$ (not necessarily) equal subsets for each processor to compute. For matrix multiplication, the most widely-used partitionings are grouped into three categories~\cite{ballard2013communication}: 1D, 2D, and 3D, based on the number of dimensions of the iteration space that are split. Figure~\ref{fig:1d2d3d} shows an example for each category. The left image is 1D partitioning ($n=8$) because only the $j$ axis is split. The middle image splits axes $i$ and $j$ so it is 2D partitioning ($n=64$). The right image is 3D partitioning ($n=64$) because all three axes are split. 

\paragraph{Owner computes.} Owner-compute strategies split a matrix (or matrices) equally among processors and each processor is responsible for all computations related to the matrix chunk it \emph{owns}. The 1D and 2D partitionings in Figure~\ref{fig:1d2d3d} are owner-compute strategies. In 1D case, each processor owns a slice of matrices $B$ and $C$ each, and computes the whole slab requires for its slice of $C$. In 2D case, each processor owns a patch of $C$ and computes a pencil corresponding to it. The 3D partitioning goes beyond owner-compute rule, since no processor is responsible for all computations associated with the data it has. Owner-compute schemes are more common because they are the most intuitive as we often view the output data as the unit of work. We will show why its communication costs are usually suboptimal in the next paragraph. 

\paragraph{Communication.} Here, we focus on the number of elements that have to be transferred by a processor. Let $V$ be the voxel subset assigned to a processor, and $V_A, V_B$, and $V_C$ be the projections of $V$ onto the $A$, $B$, and $C$ planes, respectively. The total number of elements a processor has to access to complete its computation is simply $|V_A| + |V_B| + |V_C|$, where $|\cdot|$ denotes set cardinality. Since a processor can only hold a limited amount of data in memory, the rest of the elements must come through communication. The volume of the subset designates the computational workload. As mentioned in the paper, we would like to maximize the computation-to-communication ratio, therefore we want $V$ to have as low surface-to-volume ratio as possible. Assuming $V$ only takes cuboid shapes, then the best shape is a cube. 

\emph{Owner-compute} methods fall short when it cannot partition the space into cubes. To illustrate, we compare 2D and 3D partitionings for $p=64$ processors in Figure~\ref{fig:1d2d3d}. When $u=v=w$, each pencil in the 2D partitioning has a computation-to-communication ratio,
$$r_{\text{2D}} = \dfrac{2u^3/64}{u^2/64 + 2u^2/8} = 2u/17 \approx 0.12u.$$
Each cube in the 3D partitioning has a higher computation-to-communication ratio,
$$r_{\text{3D}} = \dfrac{2u^3/64}{3u^2/16} = u/6 \approx 0.17u.$$

\paragraph{Mesh-TensorFlow.} Mesh-TensorFlow can express more fine-grained parallelism than \emph{owner-compute}, even though all we have to do is just specifying the data layout for each tensor. That is because our layout allows the tensor to be replicated. This, combines with multiple layouts from multiple tensors involved in an operation, can split the iteration space in as many dimensions as necessary (up to the rank of the iteration space). 

\input{lm1b_samples}

%% file: lm1b_samples
\section{Random Samples from Transformer Language Models Trained on Billion-Word Language Modeling Benchmark} \label{samples}

The following samples were randomly generated from the Transformer language models described in the paper.  All sentences were seeded with the initial words "According to Ray Kurzweil", and continued randomly by the model.  While all the models produce mostly grammatical sentences, the larger models exhibit more world knowledge.

\subsection{Model with 0.14B Parameters, PPL=35.0}
According to Ray Kurzweil ...
\begin{itemize}
\item ... , professor of musical economics at the University of California , Berkeley , the military 's " biggest challenge " might have been growing the region 's growing urban sprawl .
\item ... , the chief executive of ITC Emerging Markets , the metals sector should contribute about \$ 210 billion by year-end 2008 -- a figure that has now risen to \$ 3 billion .
\item ... , the head of PMI , he has now written off all his employees fairly .
\item ... , senior vice president of sales and marketing at Aspen , there is a percentage of the online response for the auto company escalating to 18 customers and increasing to 30 days a year .
\item ... , director of the Economic Policy Institute , India ranks 21st out of 221 nations in the current budget deficit , which ranks 13th globally .
\item ... , Victor Hesketh 's food agency that produces , sends and distributes the raw beef and pork through the state 's Food and Drug Administration , and they 're also allowed to take the lamb by their own account alongside the chicken and veal .
\item ... , the author of " Smoking : The Letters of Hysteria and Reclining the State of South Carolina " ( 2007 ) , 30 percent of liquor 's coconut is sold on the first batch .
\item ... , an MIT student who is not involved in anything more than a stock-market move , the latest system of debt and bankruptcy parallels the completely unregulated collection of debt that emerged in the early 1990s , when a financial industry boom brought a financial crisis that spilled almost everywhere from the United States .
\end{itemize}

\subsection{Model with 0.37B Parameters, PPL=28.9}
According to Ray Kurzweil ...
\begin{itemize}

\item ... , the owner and web guru at Stanford University , even a single person might fall in love with the internet simultaneously .
\item ... , PhD , a professor of computer science at Princeton who has staked his reputation on the ability of code technicians to simultaneously amplify a cursor 's legal weight , machines will go digital by using meta-dimensions instead of a brick , and design schemes closer to the core of GPUs .
\item ... , chief executive of the company , very few people work through chronology , and most people can 't use tables on their machines .
\item ... , we are saving the most jobs in the world .
\item ... , creator of the Star Wars creator and creator of the popular game , " Indiana Jones and the Kingdom of the Crystal Skull , " here comes Martin Schafer ( " Spider-Man 3 " ) to describe the businessman as a grand master .
\item ... , a technology expert at the Massachusetts Institute of Technology and a host of from academia , Ardipithecus ramidus was frequently incubated with solar-powered cells to frighten away and frighten out the predators , and housed in them so they could damage Lego 's control panels .
\item ... , the famously headstrong and egocentric stack-on-stack , a keyboard is more than a part of the laptop 's muscle when it is standing upright rather than crouching , and its outlet at the right end of the screen was probably the key to how the phone turned into its belly .
\item ... , founder of the Stanford-funded company , similar " recycled book " concepts are also more of an obsession , as this lets authors track their emotions in print product forms and refuse to look at one all- , -certain identity .
\end{itemize}

\subsection{Model with 1.28B Parameters, PPL=25.1}
According to Ray Kurzweil ...
\begin{itemize}

\item ... , a transplant surgeon and Harvard astrophysicist , the results are illuminated in Honolulu , Vietnam .
\item ... of UNK , a software company in California , computer games are alive and well , with millions of gamers , but most games do not make money .
\item ... , the James Watson and James Watson Professor of Physics at MIT , if we all assume that the project will continue to go ahead , we will eventually be in an era when human beings are capable of figuring out just what our genetic make-up is focused on .
\item ... , the physicist who has kept many of these principles alive , the world has vanished into " more and more of the static world " -- and in many ways we 're losing the earth 's ability to appreciate water .
\item ... , creator and the only original idea from the series , the tablet is expected to be a device that combines a USB 2.0 card with an iPad , a laptop and an iPod Touch -- with UNK sharp-wired connections to the Internet .
\item ... and a panel of experts testifying in Los Angeles , six years in Congress attempts to improve " brain " of Americans by hitting them with a \$ 50 annual fee , then having them likely pay the company \$ 3,000 for every additional year in life they are in a job .
\item ... , creator of the Review of Medical UNK , the organisation could be the " holy grail " of degenerative brain disease .
\item ... , music analyst and co-founder of zero-carbon quantum computing firm Redpoint , if you listen carefully , you 'll hear a far more universal " idea " of how this new car is supposed to be or what it should not be , or how much going to cost .
\end{itemize}

\subsection{Model with 4.9B Parameters, PPL=24.0}
According to Ray Kurzweil ...
\begin{itemize}

\item ... , chief technology officer for the US Department of Energy , aviation has " potential to be the largest and fastest growing source of consumer and commercial emissions . "
\item ... , the futurist son of the futurist who wrote The Singularity is Near , the " early days " of Google are not the right time to push forward with the next great leap .
\item ... , creator of the first modern computer , the sexy cyborg was the brainchild of an MIT professor , Thomas Harris , and a former banking entrepreneur , Henry Lee , who was looking for an UNK , a light that could be recovered and used to light up the Internet .
\item ... , the inventor of the modern personal computer , the shrinking human brain could eventually replace the Internet as a tool of human intelligence and imagination .
\item ... , the expert and co-author of " The Singularity is Near : Comprehending the Technological Future of Engineering , " people are looking for ways to protect and make their lives better .
\item ... , creator of the Gaia hypothesis , earlier computer systems should become not just more efficient , but more-efficient , increasing their efficiency by reducing human errors ( the unexpected , but often the regrettable ) and even the number of errors .
\item ... , who will make an appearance at this year 's Consumer Electronics Show in Las Vegas next week , these mobile gadgets will be able to " talk " to each other .
\item ... , the futurist turned futurist , the onset of Alzheimer 's coincided precisely with the rate of unemployment in America .
\end{itemize}